\documentclass{article}

    \usepackage[preprint]{neurips_2022}

\usepackage[utf8]{inputenc} % allow utf-8 input
\usepackage[T1]{fontenc}    % use 8-bit T1 fonts
\usepackage{hyperref}       % hyperlinks
\usepackage{url}            % simple URL typesetting
\usepackage{booktabs}       % professional-quality tables
\usepackage{amsfonts}       % blackboard math symbols
\usepackage{nicefrac}       % compact symbols for 1/2, etc.
\usepackage{microtype}      % microtypography
\usepackage{xcolor}         % colors
\usepackage{graphicx}
\usepackage{array}

\title{Task Decoding based on Eye Movements using Synthetic Data Augmentation}

\author{%
   Shanmuka Sadhu \\
   Rutgers University \\
   \texttt{shanmuka.sadhu@rutgers.edu} \\
   \And
    Arca Baran \\
    New Jersey Institute of Technology \\
   \texttt{arcabaran56@gmail.com} \\
   \And
   Preeti Pandey \\
   \texttt{preetip1529@gmail.com} \\
    \And
   Ayush Kumar \\
   SERI, Harvard Medical School \\
   \texttt{aykumar@cs.stonybrook.edu} \\
}

\begin{document}

\maketitle

\begin{abstract}
Machine learning has been extensively used in various applications related to eye-tracking research. Understanding eye movement is one of the most significant subsets of eye-tracking research that reveals the scanning pattern of an individual. Researchers have thoroughly analyzed eye movement data to understand various eye-tracking applications, such as attention mechanisms, navigational behavior, task understanding, etc. The outcome of traditional machine learning algorithms used for decoding tasks based on eye movement data has received a mixed reaction to Yarbus' claim that it is possible to decode the observer's task from their eye movements.
In this paper, to support the hypothesis by Yarbus, we are decoding tasks categories while generating synthetic data samples using well-known Synthetic Data Generators CTGAN and its variations such as CopulaGAN and Gretel AI Synthetic Data generators on available data from an in-person user study. Our results show that augmenting more eye movement data combined with additional synthetically generated improves classification accuracy even with traditional machine learning algorithms.
We see a significant improvement in task decoding accuracy from ~$28.1\%$ using Random Forest to ~$82\%$ using Inception Time when five times more data is added in addition to the $320$ real eye movement dataset sample. Our proposed framework outperforms all the available studies on this dataset because of the use of additional synthetic datasets. We validated our claim with various algorithms and combinations of real and synthetic data to show how decoding accuracy increases with the increase in the augmentation of generated data to real data.

\end{abstract}

\section{Introduction}
Visual attention is the most crucial aspect of vision science spanning various important applications and is part of everyday life. Eye movement is the driving factor in the case of visual attention, which is studied for gaze prediction, user behavior prediction, gaze guidance, etc. The importance of visual attention in understanding human behavior may be attributed to the fact that eye movements, as a part of the oculomotor system, are driven by a combination of neurological and physiological mechanisms, both voluntary and involuntary [1]. 

There has been extensive research done in grouping user behaviors based on eye movement data~[2] to find similar or distinct groups of users. Eye movements are also used as a biometric measure, making it more intriguing to explore their impact on a user's behavior or their uniqueness [1]. However, Yarbus' hypothesis [3] of predicting a user's task from their eye movements has garnered enough attention with contradictory opinions from researchers~[9][10]. Greene et al. [5] do not support the hypothesis from Yarbus, whereas other groups of researchers, such as Henderson et al. [6], Kanan et al. [7], and Borji et al. [8] support the claim in the hypothesis using more powerful machine learning algorithms. To support the claim made by Yarbus and present an additional validation, we used the similar dataset used by Tatler et al. [2]. We address this problem statement on the idea that visual task is revealed by the spatio-temporal patterns allocated during the visual attention period.

However, to further strengthen our claim and to show a significant improvement over results from researchers both in favor and in opposition of the claim, we added additional synthetically generated eye movement data for the machine learning model training. Since most eye-tracking experiments are conducted in controlled environments, in-person participants’ involvement in data collection while conducting eye-tracking research has its limitations. This also limits the volume of data collected during the experiment. Thus, it also restricts the researchers from harnessing the advancement in the field of data-intensive machine learning and deep learning algorithms in the application research of eye tracking. Decoding the category of the task from the eye movement of a participant is one such application that could be benefited a lot from having more participant data for classification using algorithms that can capture both the spatial and temporal components of the eye movement data [9].

In this paper, we have used CTGAN [12], CopulaGAN [12], and Gretel [11] API-based CTGAN algorithms to generate synthetic data from the available real eye movement dataset and five different classification algorithms (Random Forest~[19], LightGBM~[18], XGBoost~[20], HistGradBoosting~[20], and InceptionTime~[21]) to validate our claim. We tested these algorithms on various combinations of real and synthetic data, showing how task decoding accuracy increases with an increase in the augmentation of synthetically generated data to real data. Our results significantly outperform all the previously mentioned classification algorithms on the dataset used by Green et al. [5]. Thus, it is evident that task decoding is possible from eye movement data with an accuracy going as high as 82.0\%.

% The rest of the paper is organized as follows: Section 2 discusses the exploratory analysis part, where we use Data Context Map\cite{cheng2016data} to visualize task specific details in the dataset and for feature selection. Then, Section 3 discusses the proposed classifiers to classify the task on the basis of their eye movements. Section 4 presents the result in the form of a confusion matrix which contains accuracy results. Finally, Section 5 discusses the future work and gives concluding remarks.

% Our contribution -
% augmentation of eye movement data, information do not get lost, difficult to get subjects for eye tracking because most of the experiments are controlled. It remains consistent and infact gets better and better.
\section{Methodology}
We used the eye movement dataset used by Tatler et al. [4] who replicated the previous by Yarbus. For better decoding accuracy and to strengthen the claim, we added synthetically generated data to real data. We used various classifiers to decode the task based on eye movement data by combining both real and synthetic data. Details are provided in the below subsections: 1) Details on datasets and their nature, 2) algorithms used to generate synthetic data, and 3) classification algorithms used for decoding the task from eye movement data.

\subsection{Dataset and Task}
To validate our claim, we used the eye movement dataset (anonymous) from Tatler et al. [4] which was recorded on images from the LIFE archive on Google (http://images.google.com/hosted/life). This dataset contains four tasks performed by 16 participants on 20 grayscale images. Each user looked at 20 images while performing four different tasks, which makes the dataset constituted of 320 real eye movement data samples in total, with 80 samples for each task. $Task 1$ represents the action of determining the~\textit{Decade} in which an image was taken; $Task 2$ represents the action of~\textit{memorizing} a picture; $Task 3$ represents the action of determining how well you know the~\textit{people} in a picture; and $Task 4$ represents the action of determining the~\textit{wealth} of the people in a picture. %We have shown four grayscale images as the samples from Google's LIFE archive used as stimuli for this experiment in Appendix A.1. 
We considered four features of eye movement data: 1) x-coordinate, 2) y-coordinate, 3) fixation duration, and 4) pupil diameter for training and testing purposes during the decoding. The dataset is from the right eye only, recorded at a frequency of 1000Hz.

\subsection{Synthetic Data Generator}
To generate synthetic data to be used for data augmentation while training the machine learning model, we used the widely popular CTGAN architecture and its variants CopulaGAN in this paper.

CTGAN is one of the widely-used generative models~[13] that learns the distribution of tabular data and sample rows with the distribution. It can mix discrete and continuous data through Gumbel Softmax and tanh and address the issue with continuous values, which causes the vanishing gradient problem.

CopulaGAN is another generative adversarial network used to create synthetic data. It is similar to CTGAN but uses CDFs (Cumulative Distribution Functions) from GaussianCopula~[16] to increase the generator’s effectiveness in learning the training dataset. The GaussianCopula model, based on copula distributional functions [10], is another generative model that CopulaGAN relies on CDFs.

For better performance and to avoid hyperparameter tuning (which is one of the biggest bottlenecks for GAN architecture), we used the CTGAN algorithm based on Gretel.ai's API [11] to generate time series synthetic data, which outperforms both CTGAN and CopulaGAN in terms of the quality of data it produces.

\subsubsection{Kolmogorov–Smirnov (KS) test}
We used the Kolmogorov-Smirnov test, or KS test, to assess the differences between the distributions of the synthetic data and the original eye-tracking data. The null hypothesis of the test is that the two sample data sources have identical cumulative distribution functions. The statistical test is executed on all the compatible columns (four columns in our dataset). The KS test metric we used in our paper from the Synthetic Data Vault (SDV) [14] uses the two-sample Kolmogorov–Smirnov test to compare the distributions of continuous columns using the empirical CDF.

% The Kolmogorov-Smirnov (KS) test is a nonparametric statistical test that is used to compare the empirical cumulative distribution functions (CDFs) of two samples. It can be used to assess whether two samples come from the same underlying distribution or whether there are significant differences between the two samples. The null hypothesis of the KS test is that the two samples come from the same distribution, and the test statistic is the maximum difference between the two CDFs. If the difference between the two CDFs is large, this suggests that the two samples come from different distributions and the null hypothesis can be rejected.

% In our study, we used the KS test to compare the distributions of synthetic eye-tracking data and real eye-tracking data. We applied the test to all four columns of the eye-tracking data (x-coordinate, y-coordinate, fixation duration, and pupil diameter) to determine whether the synthetic data was similar to the real data in terms of its statistical properties. We used the two-sample KS test implemented in the Synthetic Data Vault (SDV) [14], which compares the distributions of continuous columns using the empirical CDF. This allowed us to assess whether the synthetic data was a good representation of the real data and whether it could be used to supplement the real data for training machine learning models.

\subsection{Task decoders}
Over the course of the experiment, we used five different sets of algorithms to assess the generalization of the dataset as a whole with varying amounts of synthetic data in combination with real data. We have used Random Forest (RF) [19], LightGBM (LGBM) [18], Gradient Boosting (GB) [20], HistGradientBoosting(HGB) [20], and InceptionTime Classifier (ITC) [21]. We used the InceptionTimePlus variation of the InceptionTime classifier algorithm from a state-of-the-art deep learning library for time series classification [22]. It is an ensemble of deep Convolutional Neural Network (CNN) models and five deep learning models for time series classification, each one created by cascading multiple Inception modules [23], inspired by the Inception-v4 architecture [24]. Each classifier (model) has the same architecture but different randomly initialized weight values.

\section{Results}

\subsection{Quality of Data generated}
To generate the synthetic tabular data, we used CTGAN, one of the widely used algorithms. We used the SDV libraries, which allow users to generate new synthetic time series tabular data with the same format and statistical properties as the original dataset. However, the KS Test for the data generated from CTGAN only had a score of $0.73$. We also used the CopulaGAN variation of CTGAN from the SDV libraries, which had a KS Test score of $0.83$. To generate a more reliable and statistically similar dataset to the real dataset, we used CTGAN available from Gretel.ai (C-CTGAN) based API, which resulted in a KS test score of $0.9$. For illustrative purposes, we have used Figure 2 in Appendix A as a qualitative performance measure of the synthetic data generated using these 3 algorithms. It is evident from the images as well that the dataset produced by C-CTGAN mimics the real data very well, whereas CopulaGAN smooths it out, and CTGAN also fails to capture the spatial spread of the fixations.

\subsection{Task decoding results}

   \begin{table}[h!]
    \begin{tabular}{|c|c|c|c|c|c|c|} 
     \hline
      Data Comb & Algo & RF & LGBM & GB & HGB & ITC \\ [0.5 ex] 
     \hline
      
     320R & & $28.1 \pm 0.39$ & $32.8 \pm 0.01$ & $35.9 \pm 0.3$ & $34.4 \pm 0.01$ & $34.6 \pm 0.14$ \\ 
     		
     \hline
     \\ 
     320R + 960S & CTGAN & $34.4 \pm 0.25$ & $34.7 \pm 0.01$ & $37.1 \pm 0.31$ & $40.9 \pm 0.01$ & $\textbf{73.0} \pm 0.22$ \\ 
    \\ 
     320R + 1600S & CTGAN & $34.6 \pm 0.12$ & $37.0 \pm 0.01$ & $38.8 \pm 0.32$ & $41.1 \pm 0.01$ & $75.0\pm 0.22$ \\ 
     \hline
     \\ 
     320R + 960S & CopulaGAN & $38.4 \pm 0.35$ & $37.9 \pm 0.01$ & $37.9 \pm 0.22$ & $42.2 \pm 0.01$ & $71.4 \pm 0.16$ \\ 
     \\ 
     320R + 1600S & CopulaGAN & $44.5 \pm 0.15$ & $39.1 \pm 0.01$ & $38.3 \pm 0.22$ & $43.8 \pm 0.01$ & $74.7 \pm 0.19$ \\ 
    \hline
    \\ 
     320R + 960S & G-CTGAN & $47.3 \pm 0.06$ & $56.3 \pm 0.01$ & $52.7 \pm 0.07$ & $54.7 \pm 0.01$ & $70.0 \pm 0.07$ \\ 
     \\ 
     320R + 1600S & G-CTGAN& $66.9 \pm 0.23$ & $67.4 \pm 0.01$ & $65.9 \pm 0.17$ & $65.6 \pm 0.01$ & $\textbf{82.0} \pm 0.18$ \\ 
    \hline
    \end{tabular}
       \\
    \caption{Accuracies along with standard deviation using all five models with $320R$, $320R + 960S$ and $320R + 1600S$ respectively using CTGAN, CopulaGAN, and C-CTGAN} 
    \end{table} 
 
To validate that using powerful machine learning algorithms will positively impact the possibility of task decoding, we trained multiple powerful algorithms on real datasets, and various combinations of the synthetically generated dataset from the three previously mentioned synthetic data generation approaches. The dataset was split into 80 \% train and 20 \% holdout test sets. We ensured the training set remained equally divided into four parts local training sets and the holdout test case. All the accuracies reported in this paper are the average performance over five repetitions of the experiment. We also report the standard deviation over the repetitions as confidence bounds with respective average accuracies, as shown in Table 1.

We tried using five different combinations of real and synthetic data for our model training ($80\%$) and testing ($20\%$) during task decoding. We trained and tested our model on 1) 320 samples of real data ($320R$), 2) 320 real samples + 320 synthetic data samples ($320R + 320S$), 3) 320 real samples + 640 synthetic data samples ($320R + 640S$), 4) 320 real samples + 960 synthetic data samples ($320R + 960S$), 5) 320 real samples + 1600 synthetic data samples ($320R + 1600S$). With only 320 real data samples, even selecting powerful machine learning algorithms did not yield very good results and went as high as 35.9\% using gradient boosting from 28.1\% using random forest. Even ITC gave an accuracy of 34.6\%. Since real data samples alone were insufficient for the model to learn the features properly, we augmented our model with an additional $320S$, $640S$, $960S$, and $1600S$ samples of data, respectively. Additional $1600S$ augmented data improve the decoding accuracy subsequently for all the algorithms ranging from 65.6\% using HGB to 82.0\% using ITC, as shown in Table 1. Augmenting synthetic data generated from G-CTGAN and decoding performance using ITC, respectively, outperform overall, as shown in Table 1. It was also evident from the quality of data generated by three different versions of CTGAN algorithms, as discussed in Section 3.1.

We also evaluated the performance using a combination of $320R + 320S$ and $320R + 640S$ and tabulated the result in Table 2 of Appendix A for reference.

\section{Discussion and Conclusion}
\label{conclusion}
In this paper, we aim to support the claim that it is possible to decode an individual task with the help of eye movements associated with them. We follow this claim based on findings that eye gazes from eye movement data carry cognitive information such as a mental state which is highly related to the task the observer is carrying out. Using multiple machine learning-based decoding algorithms, we can predict the observer’s task from eye movement data. While our experiment yields better than most of the previously mentioned results for a similar task and set of data, we additionally augmented synthetic data to real data samples for better decoding accuracy and robustness. We witnessed a significant improvement in accuracy ranging from as low as ~28.1\% using Random Forest on real data samples to ~82\% using Inception Time when adding five times more data in addition to the 320 real dataset samples.

Even though we have additional eye movement features, such as saccades, blinks, etc., available, our paper does not use these features for evaluation purposes. We are limited primarily to fixation duration, x - y coordinate, and pupil size. In future work, we plan to consider all available features for more robust decoding using all possible behavioral cues. Using all possible available eye movement features and the additional behavioral cues might also open the research paradigm, which helps us to possibly identify the individual based on their gaze behavior. This will help us identify the uniqueness in each individual visual attention and navigational pattern and will enable researchers to push the research further to consider eye movement biometrics as a level of authentication performance acceptable for real-world use.

\section*{References}
{
\small

[1] Lohr, D., \& Komogortsev, O. V. (2022). Eye Know You Too: Toward Viable End-to-End Eye Movement Biometrics for User Authentication. {\it IEEE Transactions on Information Forensics and Security.}

[2] Kumar, A., Netzel, R., Burch, M., Weiskopf, D., \& Mueller, K. (2017). Visual multi-metric grouping of eye-tracking data. {\it Journal of Eye Movement Research}, 10(5).

[3] Yarbus, A. L. (1967). {\it Eye movements and vision}. Springer.

[4] Tatler, B. W., Wade, N. J., Kwan, H., Findlay, J. M., \& Velichkovsky, B. M. (2010). Yarbus, eye movements, and vision. {\it i-Perception,} 1(1), 7-27.

[5] Greene, M. R., Liu, T., \& Wolfe, J. M. (2012). Reconsidering Yarbus: A failure to predict observers’ task from eye movement patterns. {\it Vision research}, 62, 1-8.

[6] Henderson, J. M., Shinkareva, S. V., Wang, J., Luke, S. G., \& Olejarczyk, J. (2013). Predicting cognitive state from eye movements.{\it PloS one}, 8(5), e64937.

[7] Kanan, C., Ray, N. A., Bseiso, D. N., Hsiao, J. H., \& Cottrell, G. W. (2014, March). Predicting an observer's task using multi-fixation pattern analysis. In {\it Proceedings of the symposium on eye tracking research and applications} (pp. 287-290).

[8] Borji, A., \& Itti, L. (2014). Defending Yarbus: Eye movements reveal observers' task. {\it Journal of vision}, 14(3), 29-29.

[9] Kumar, A., Tyagi, A., Burch, M., Weiskopf, D., \& Mueller, K. (2019). Task classification model for visual fixation, exploration, and search. In {\it Proceedings of the 11th ACM Symposium on Eye Tracking Research \& Applications} (pp. 1-4).

[10] Kumar, A., Howlader, P., Garcia, R., Weiskopf, D., \& Mueller, K. (2020). Challenges in interpretability of neural networks for eye movement data. In {\it ACM Symposium on Eye Tracking Research and Applications} (pp. 1-5).

[11] Gretel.Ai - privacy engineering as a service (2022).  {\it Gretel.ai. [Online]}. Available: https://gretel.ai/

[12] Xu, L., Skoularidou, M., Cuesta-Infante, A., \& Veeramachaneni, K. (2019). Modeling tabular data using conditional gan. {\it Advances in Neural Information Processing Systems}, 32.

[13] Goodfellow, I., Pouget-Abadie, J., Mirza, M., Xu, B., Warde-Farley, D., Ozair, S., ... \& Bengio, Y. (2020). Generative adversarial networks. {\it Communications of the ACM}, 63(11), 139-144.

[14] Patki, N., Wedge, R., \& Veeramachaneni, K. (2016). The synthetic data vault. In 2016 {\it IEEE International Conference on Data Science and Advanced Analytics (DSAA)} (pp. 399-410). IEEE.

[15] Eberle, O., Brandl, S., Pilot, J., \& Søgaard, A. (2022). Do Transformer Models Show Similar Attention Patterns to Task-Specific Human Gaze?. {\it In Proceedings of the 60th Annual Meeting of the Association for Computational Linguistics (Volume 1: Long Papers)} (pp. 4295-4309).

[16] Nelsen, R. B. (2007). An introduction to copulas. {\it Springer Science \& Business Media}.

[17] Pedregosa, F., Varoquaux, G., Gramfort, A., Michel, V., Thirion, B., Grisel, O., ... \& Duchesnay, E. (2011). Scikit-learn: Machine learning in Python.  {\it the Journal of machine Learning research}, 12, 2825-2830.

[18] Ke, G., Meng, Q., Finley, T., Wang, T., Chen, W., Ma, W., ... \& Liu, T. Y. (2017). Lightgbm: A highly efficient gradient boosting decision tree. {\it Advances in neural information processing systems}, 30.

[19] Liaw, A., \& Wiener, M. (2002). Classification and regression by randomForest.  {\it R news}, 2(3), 18-22.

[20] Polikar, R. (2012). Ensemble learning. In  {\it Ensemble machine learning} (pp. 1-34). Springer, Boston, MA.

[21] Ismail Fawaz, H., Lucas, B., Forestier, G., Pelletier, C., Schmidt, D. F., Weber, J., ... \& Petitjean, F. (2020). Inceptiontime: Finding alexnet for time series classification. {\it Data Mining and Knowledge Discovery}, 34(6), 1936-1962.

[22] Oguiza, I. (2022). tsai-A state-of-the-art deep learning library for time series and sequential data. {\it Retrieved September}, 19, 2022.

[23] Szegedy, C., Liu, W., Jia, Y., Sermanet, P., Reed, S., Anguelov, D., ... \& Rabinovich, A. (2015). Going deeper with convolutions. In {\it Proceedings of the IEEE conference on computer vision and pattern recognition} (pp. 1-9).

[24] Szegedy, C., Ioffe, S., Vanhoucke, V., \& Alemi, A. A. (2017). Inception-v4, inception-resnet and the impact of residual connections on learning. In {\it Thirty-first AAAI conference on artificial intelligence}.

}

%%%%%%%%%%%%%%%%%%%%%%%%%%%%%%%%%%%%%%%%%%%%%%%%%%%%%%%%%%%%
\section*{Checklist}

%%% BEGIN INSTRUCTIONS %%%
% The checklist follows the references.  Please
% read the checklist guidelines carefully for information on how to answer these
% questions.  For each question, change the default \answerTODO{} to \answerYes{},
% \answerNo{}, or \answerNA{}.  You are strongly encouraged to include a {\bf
% justification to your answer}, either by referencing the appropriate section of
% your paper or providing a brief inline description.  For example:
% \begin{itemize}
%   \item Did you include the license to the code and datasets? \answerYes{See Section~\ref{gen_inst}.}
%   \item Did you include the license to the code and datasets? \answerNo{The code and the data are proprietary.}
%   \item Did you include the license to the code and datasets? \answerNA{}
% \end{itemize}
% Please do not modify the questions and only use the provided macros for your
% answers.  Note that the Checklist section does not count towards the page
% limit.  In your paper, please delete this instructions block and only keep the
% Checklist section heading above along with the questions/answers below.
%%% END INSTRUCTIONS %%%

\begin{enumerate}

\item For all authors...
\begin{enumerate}
  \item Do the main claims made in the abstract and introduction accurately reflect the paper's contributions and scope?
    \answerYes{}
  \item Did you describe the limitations of your work?
    \answerYes{}
  \item Did you discuss any potential negative societal impacts of your work?
    \answerNo{}
  \item Have you read the ethics review guidelines and ensured that your paper conforms to them?
    \answerNA{}
\end{enumerate}

\item If you are including theoretical results...
\begin{enumerate}
  \item Did you state the full set of assumptions of all theoretical results?
    \answerNA{}
        \item Did you include complete proofs of all theoretical results?
    \answerNA{}
\end{enumerate}

\item If you ran experiments...
\begin{enumerate}
  \item Did you include the code, data, and instructions needed to reproduce the main experimental results (either in the supplemental material or as a URL)?
    \answerNo{Can be provided if asked}
  \item Did you specify all the training details (e.g., data splits, hyperparameters, how they were chosen)?
    \answerYes{}
        \item Did you report error bars (e.g., with respect to the random seed after running experiments multiple times)?
    \answerYes{As standard deviation}
        \item Did you include the total amount of compute and the type of resources used (e.g., type of GPUs, internal cluster, or cloud provider)?
    \answerNo{}
\end{enumerate}

\item If you are using existing assets (e.g., code, data, models) or curating/releasing new assets...
\begin{enumerate}
  \item If your work uses existing assets, did you cite the creators?
    \answerYes{}
  \item Did you mention the license of the assets?
    \answerNo{}
  \item Did you include any new assets either in the supplemental material or as a URL?
    \answerNo{}
  \item Did you discuss whether and how consent was obtained from people whose data you're using/curating?
    \answerNo{We would acknowledge at the end after accepted}
  \item Did you discuss whether the data you are using/curating contains personally identifiable information or offensive content?
    \answerYes{}
\end{enumerate}

\item If you used crowdsourcing or conducted research with human subjects...
\begin{enumerate}
  \item Did you include the full text of instructions given to participants and screenshots, if applicable?
    \answerNA{}
  \item Did you describe any potential participant risks, with links to Institutional Review Board (IRB) approvals, if applicable?
    \answerNA{}
  \item Did you include the estimated hourly wage paid to participants and the total amount spent on participant compensation?
    \answerNA{}
\end{enumerate}

\end{enumerate}

%%%%%%%%%%%%%%%%%%%%%%%%%%%%%%%%%%%%%%%%%%%%%%%%%%%%%%%%%%%%

\appendix

\section{Appendix}

% \subsection{Dataset and Task}

\begin{figure}
  \centering
%   \fbox{\rule[-.5cm]{0cm}{4cm} \rule[-.5cm]{4cm}{0cm}}
  \includegraphics[height=2in, width=2.7in]{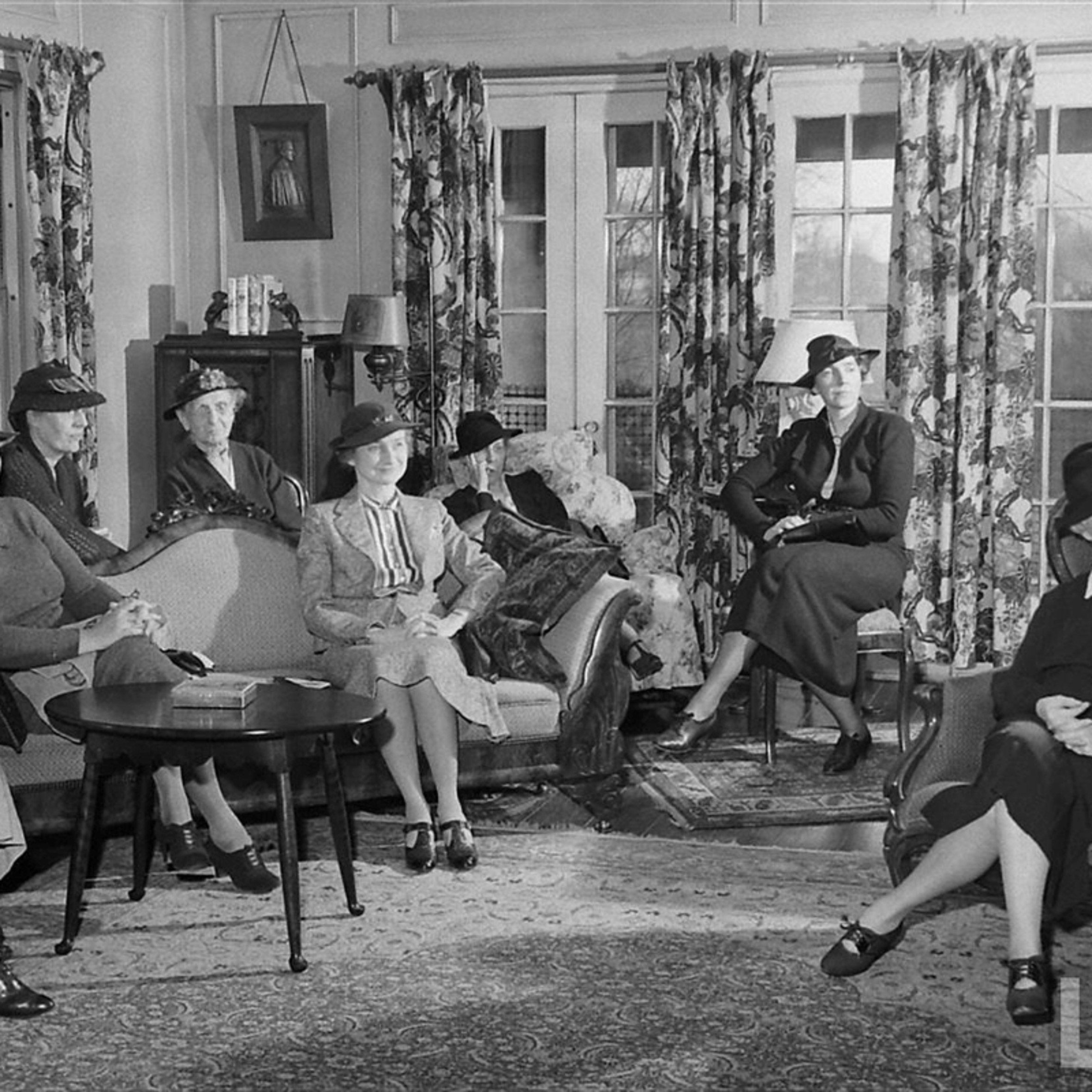}
  \includegraphics[height=2in, width=2.7in]{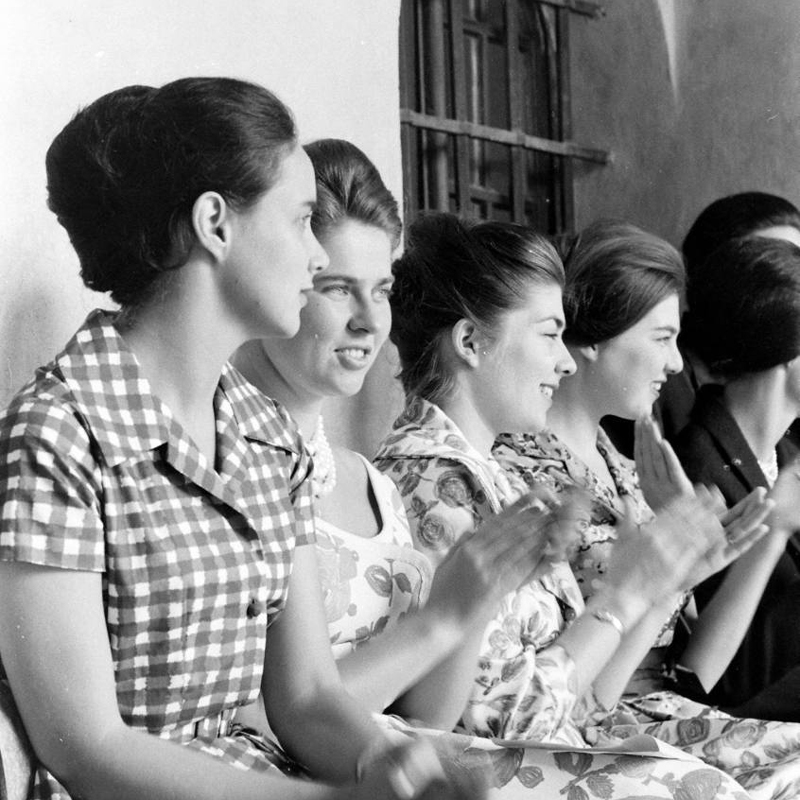}\\
  \includegraphics[height=2in, width=2.7in]{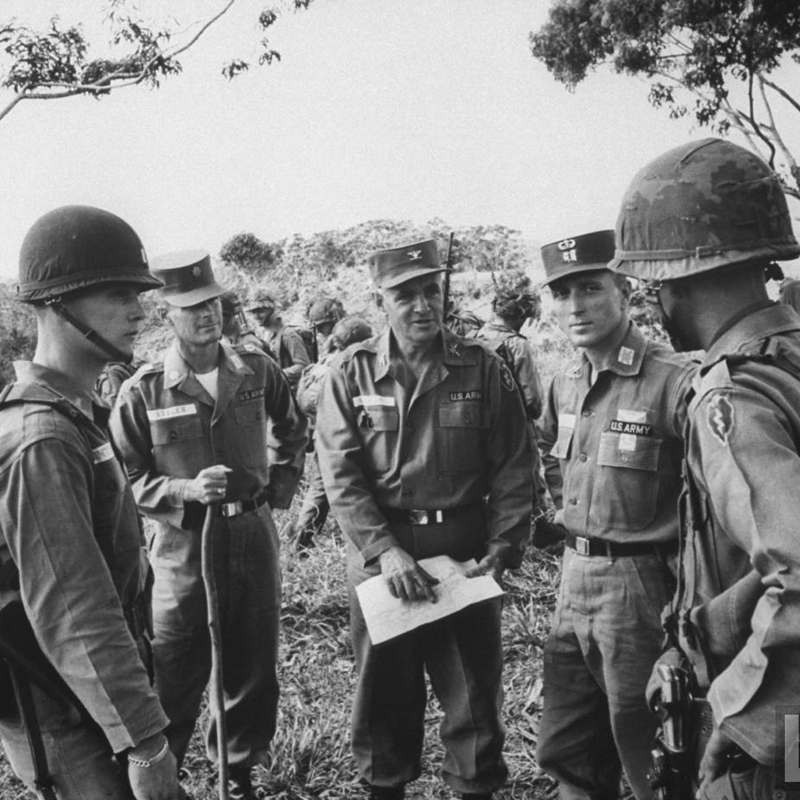}
  \includegraphics[height=2in, width=2.7in]{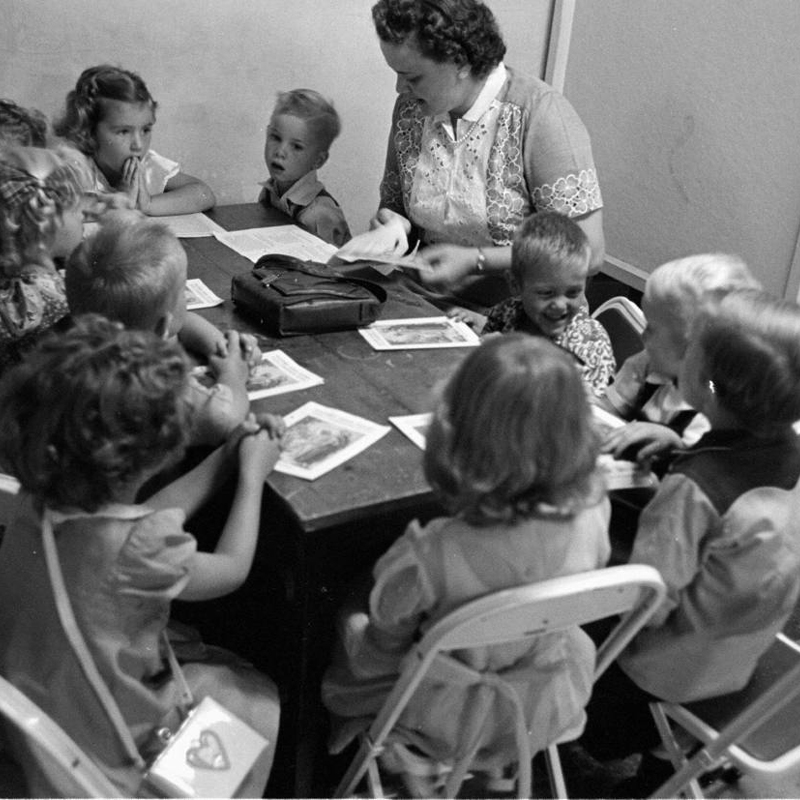}\\
   
  \caption{These four gray scales images are the sample from Time Life archive on Google (http://images.google.com/hosted/life) and is used as stimuli for this experiment which consists of 20 grayscale images in total}
\end{figure}

% \subsection{Synthetic Data Generator}

\begin{figure}[h!]
  \centering
  $\begin{array}{cc}
%   \fbox{\rule[-.5cm]{0cm}{4cm} \rule[-.5cm]{4cm}{0cm}}
   \includegraphics[height=2in, width=2.72in]{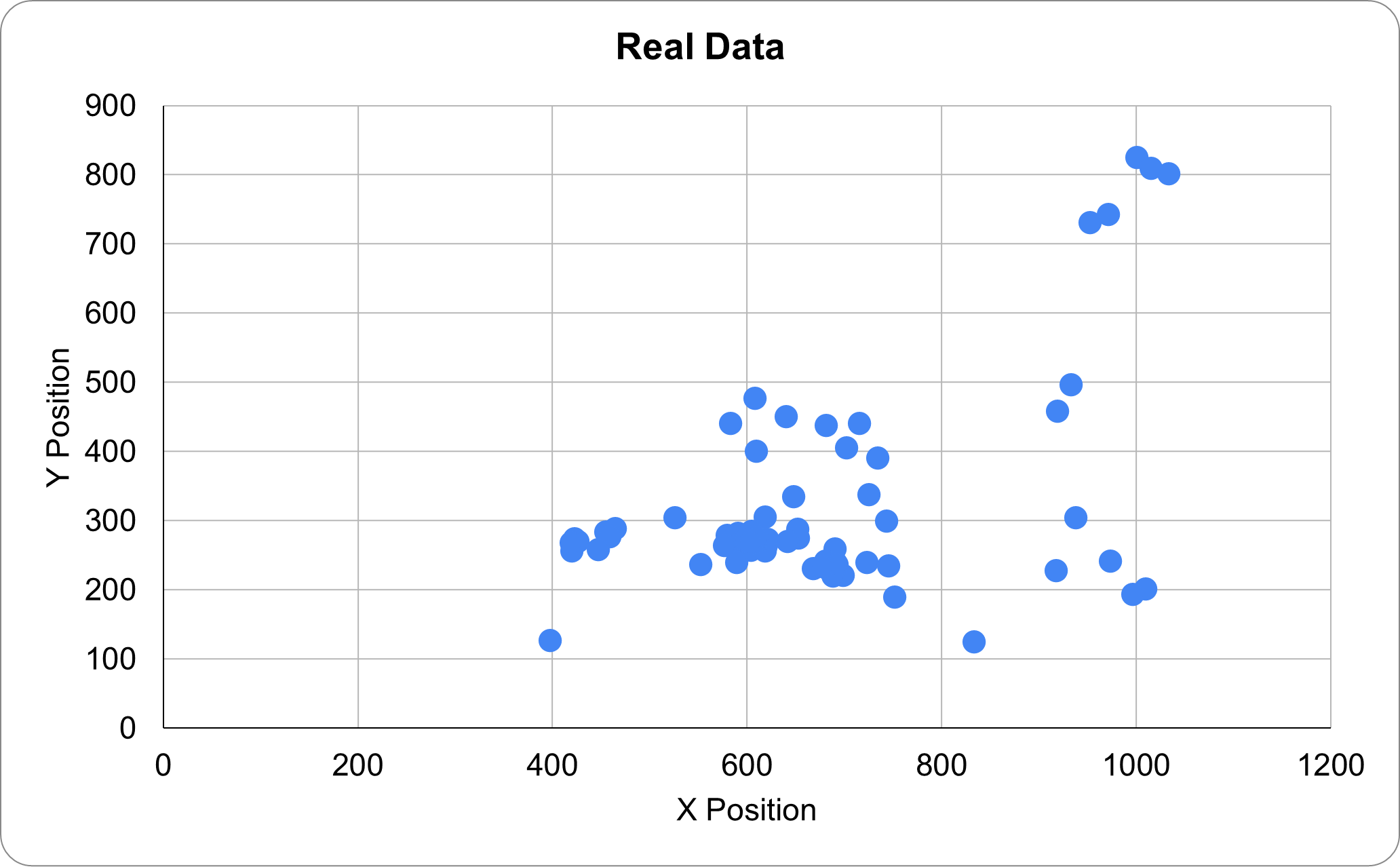} &
     \includegraphics[height=2in, width=2.72in]{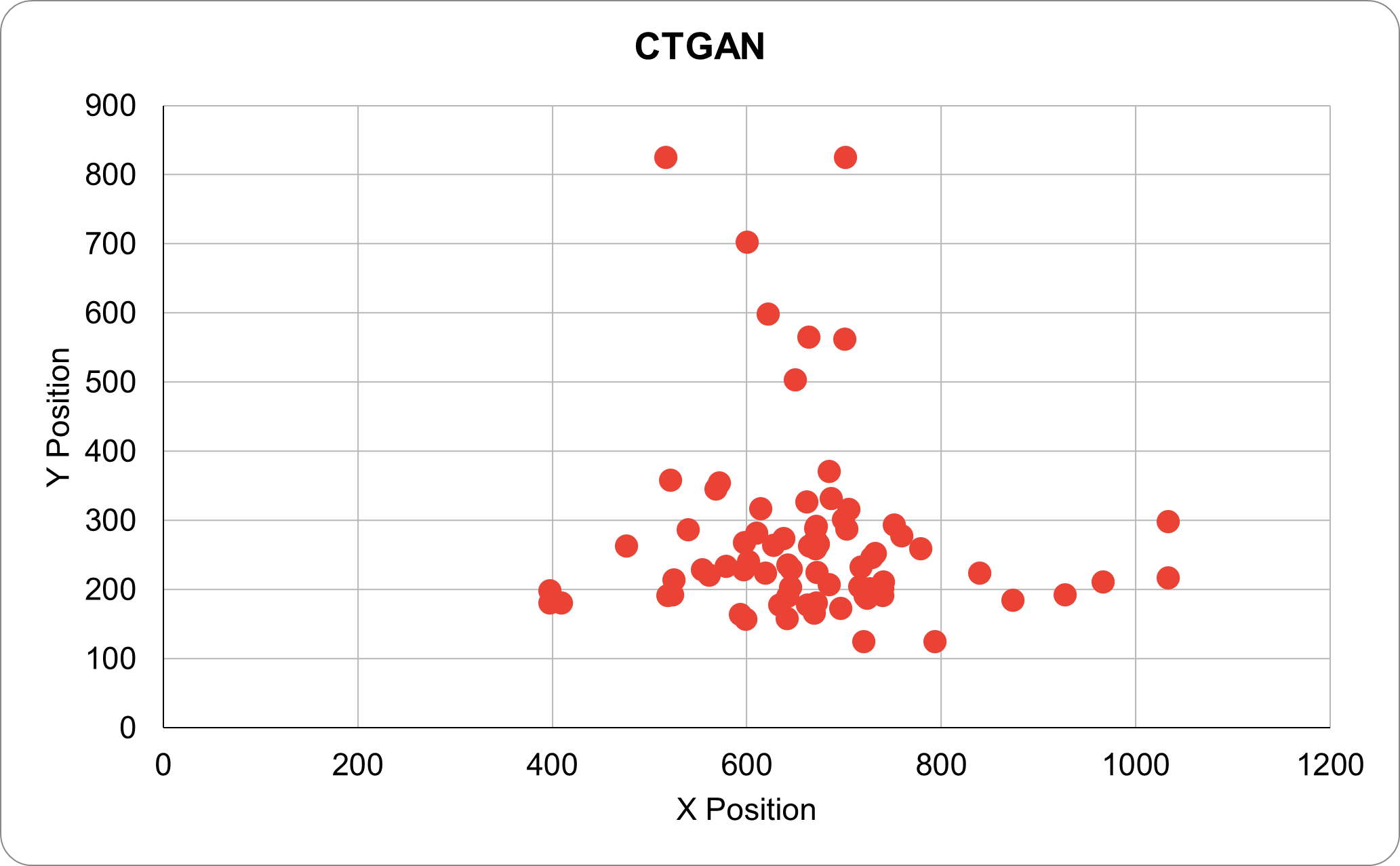} \\
      \mbox{(a)} & \mbox{(b)}\\
      \includegraphics[height=2in, width=2.72in]{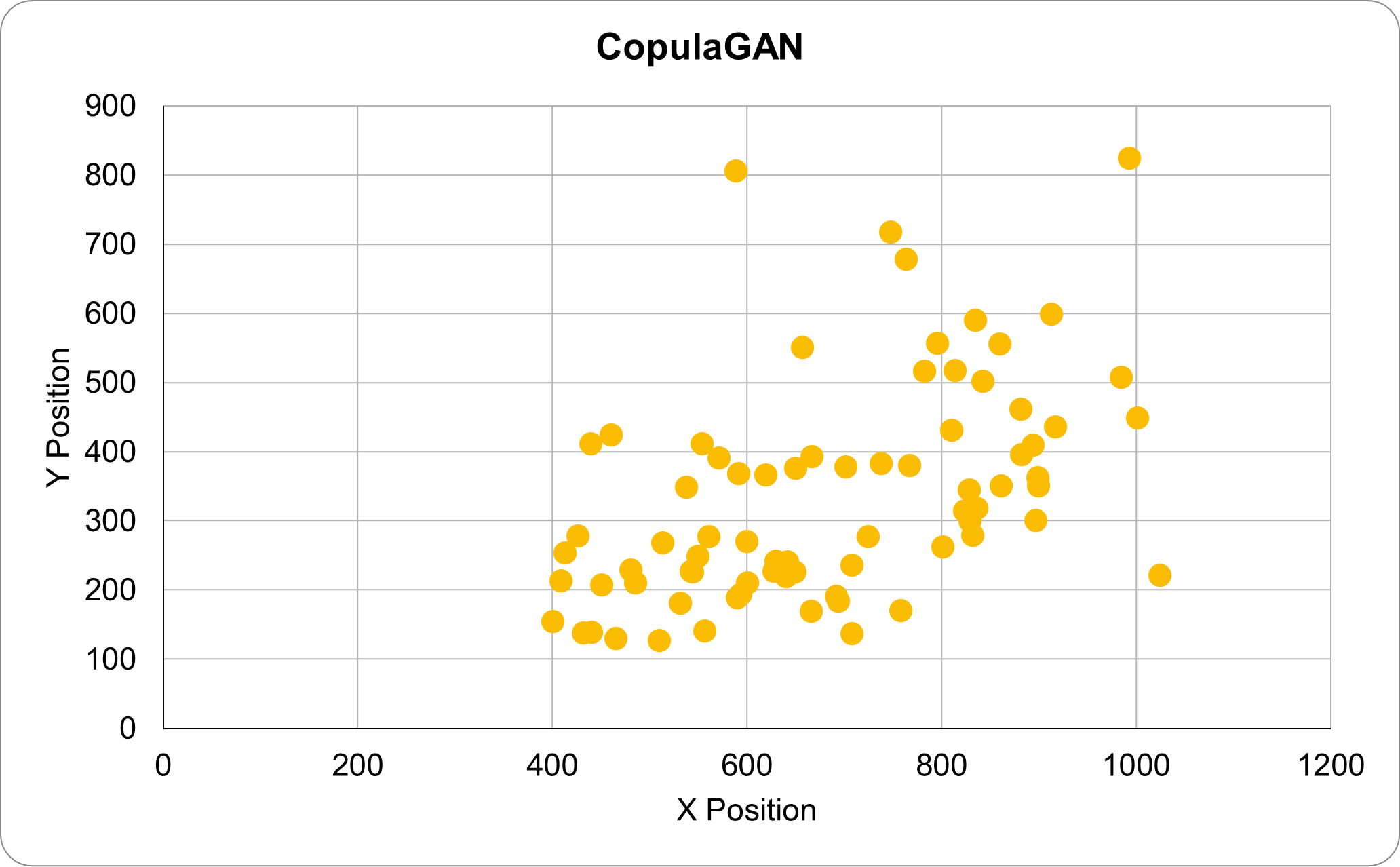} &
     \includegraphics[height=2in, width=2.72in]{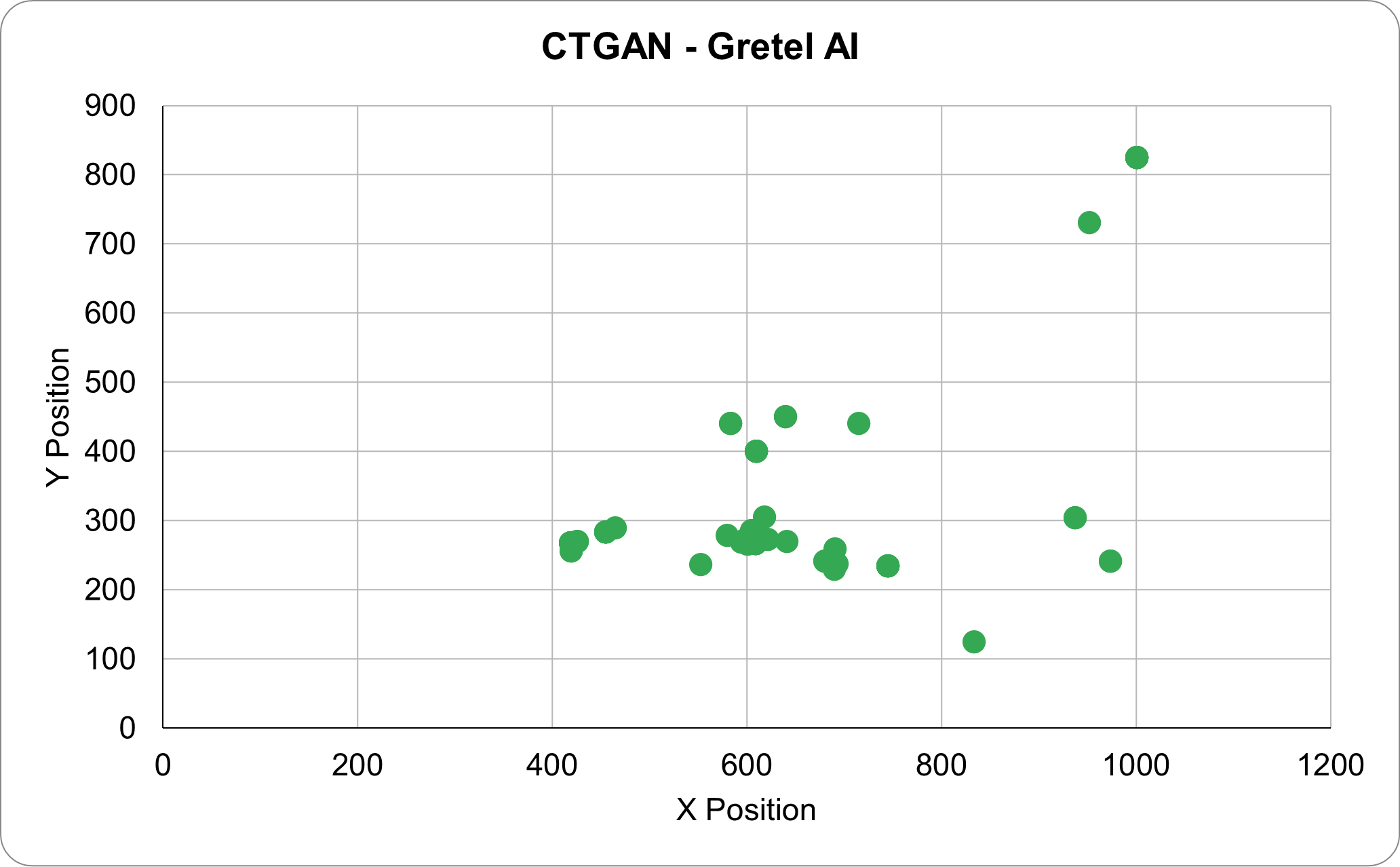} \\
      \mbox{(c)} & \mbox{(d)}
    \end{array}$
  \caption{A sample plot of real data collected via in-person experiment as shown in (a). Whereas (b), (c) and (d) are the synthetic data generated from algorithms such as CTGAN, CopulaGAN, and CTGAN from GretelAI api (C-CTGAN), respectively. Synthetic data generated using GretelAI based api using CTGAN in background produces nearly identical data in spatial context.}
\end{figure}

% \subsection{Results}

   \begin{table}[h!]
    \begin{tabular}{|c|c|c|c|c|c|c|} 
     \hline
      Data Combination & Algo & RF & LGBM & GB & HGB & ITC \\ [0.5 ex] 
     \hline
      
          320R & & $28.1 \pm 0.39$ & $32.8 \pm 0.01$ & $35.9 \pm 0.3$ & $34.4 \pm 0.01$ & $34.6 \pm 0.14$ \\  
     		
     \hline
     \\ 
     320R + 320S & CTGAN & $32.8 \pm 0.14$  & $32.3 \pm 0.01$ & $43.0 \pm 0.21$ & $39.1 \pm 0.01$ & $70.0 \pm 0.24$ \\ 
    \\ 
     320R + 640S & CTGAN & $33.3 \pm 0.09$  & $33.6 \pm 0.01$ & $48.0 \pm 0.13$ & $38.7 \pm 0.01$ & $64.0 \pm 0.20$ \\ 
     \hline
     \\ 
     320R + 320S & CopulaGAN & $35.9 \pm 0.40$  & $31.0 \pm 0.01$ & $35.9 \pm 0.20$ & $38.3 \pm 0.01$ & $75.2 \pm 0.28$ \\ 
     \\ 
     320R + 640S & CopulaGAN & $34.3 \pm 0.20$  & $32.8 \pm 0.01$ & $36.7 \pm 0.19$ & $40.0 \pm 0.01$ & $73.0 \pm 0.12$ \\ 
    \hline
    \\ 
     320R + 320S & G-CTGAN & $44.5 \pm 0.29$  & $46.1 \pm 0.01$ & $43.0 \pm 0.32$ & $50.0 \pm 0.01$ & $70.7 \pm 0.11$ \\ 
     \\ 
     320R + 640S & G-CTGAN & $44.3 \pm 0.21$  & $53.1 \pm 0.01$ & $48.0\pm 0.07$ & $49.5 \pm 0.01$ & $73.8 \pm 0.14$ \\ 
    \hline
    \end{tabular}
       \\
    \caption{Accuracy using all five models with $320R + 320S$ and $320R + 640S$ respectively using CTGAN, CopulaGAN, and C-CTGAN.}
    \end{table} 

\end{document}